\documentclass{article} 
\usepackage{colm2024_conference}

\usepackage{microtype}
\usepackage{hyperref}
\usepackage{url}
\usepackage{booktabs}
\usepackage{graphicx}
\usepackage{setspace}
\usepackage{listings}
\usepackage{algorithm}
\usepackage{algorithmic}
\usepackage{amsmath}

\definecolor{correct_response_color}{RGB}{0,176,80}
\definecolor{wrong_response_color}{RGB}{192,0,0}
\definecolor{idk_response_color}{RGB}{0,112,192}

\newcommand{\piref}{\pi_\text{ref}}

\title{Can AI Assistants Know What They Don't Know?}

\colmfinalcopy

\author{
Qinyuan Cheng$^{1,2,}$\thanks{Equal contribution.} \hspace{.3em}
Tianxiang Sun$^{1,2,}$\protect\footnotemark[\value{footnote}] \hspace{.3em}
Xiangyang Liu$^{1,2}$ \hspace{.1em}
Wenwei Zhang$^{2}$
\\
\textbf{
Zhangyue Yin$^{1}$ \hspace{.1em}
Shimin Li$^{1}$ \hspace{.1em}
Linyang Li$^{1,2}$ \hspace{.1em}
Zhengfu He$^{1}$ \hspace{.1em}
}
\\
\textbf{
Kai Chen$^{2,}$\thanks{Corresponding author. 
 Correspondence to: Qinyuan Cheng \texttt{<chengqy2019@foxmail.com>} Kai Chen \texttt{<chenkai@pjlab.org.cn>} Xipeng Qiu \texttt{<xpqiu@fudan.edu.cn>}} \hspace{.2em}
Xipeng Qiu$^{1,}$\protect\footnotemark[\value{footnote}]
}
\\
[1ex]
$^{1}$Fudan University \\
$^{2}$Shanghai AI Laboratory \\
}

\fancypagestyle{firstpage}{
  \lhead{\begin{picture}(0,0)\put(0,-6){\includegraphics[width=0.3\linewidth]{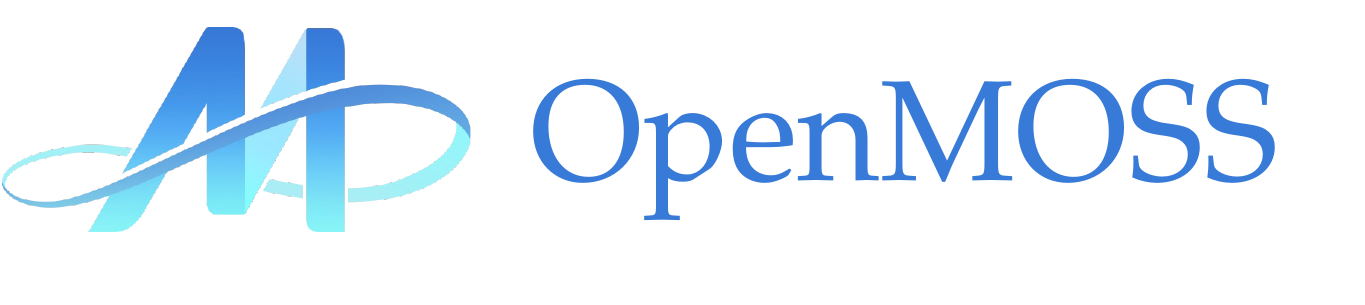}}\end{picture}}
}

\begin{document}

\maketitle

\thispagestyle{firstpage}

\begin{abstract}
Recently, AI assistants based on large language models (LLMs) show surprising performance in many tasks, such as dialogue, solving math problems, writing code, and using tools.
Although LLMs possess intensive world knowledge, they still make factual errors when facing some knowledge intensive tasks, like open-domain question answering.
These untruthful responses from the AI assistant may cause significant risks in practical applications.
We believe that an AI assistant's refusal to answer questions it does not know is a crucial method for reducing hallucinations and making the assistant truthful.
Therefore, in this paper, we ask the question ``\textbf{Can AI assistants know what they don't know and express them through natural language?}"
To answer this question, we construct a model-specific ``I don't know" (Idk) dataset for an assistant, which contains its known and unknown questions, based on existing open-domain question answering datasets.
Then we align the assistant with its corresponding Idk dataset and observe whether it can refuse to answer its unknown questions after alignment.
Experimental results show that after alignment with Idk datasets, the assistant can refuse to answer most its unknown questions.
For questions they attempt to answer, the accuracy is significantly higher than before the alignment.
\footnote{We will release our code, data and models at \url{https://github.com/OpenMOSS/Say-I-Dont-Know}.}
\end{abstract}




\section{Introduction}
Large language models \citep{GPT-3, PaLM, GLM130B, llama1} possess extensive world knowledge and demonstrate capabilities in numerous natural language tasks that smaller models lack \citep{Emergent_Abilities}.
Recently, there are many artifical intelligence chat assistants built on large language models, capable of helping users accomplish various tasks in the daily life, providing a satisfactory user experience \citep{InstructGPT, ChatGPT, Claude, sun2023moss, baichuan2, Qwen}.
Although these chat assistants can communicate frequently with users, they are prone to hallucinations \citep{Hallucination_in_Conversation, Hallucination_survey, HalluQA}, such as including factual errors in their generated responses \citep{Factuality_Survey} or imitating human falsehoods in training corpus \citep{TruthfulQA}, some of which are difficult for users to detect.
These untruthful responses could potentially harm society and also diminish the credibility of AI assistants.

An AI assistant aligned with human values should be truthful \citep{TruthfulAI}, which means that it needs to provide accurate information consistent with the real world.
When the assistant's output contains factual errors, it indicates that it may lack the corresponding knowledge internally, yet it fails to express the unknowns and refuse to give a answer.
However, a truthful AI assistant should be aware of what it knows and what it does not know, and be able to communicate this to the user.
For questions that are known, the AI assistant should provide users with accurate information.
For questions that are unknown, the AI assistant should avoid giving answers.
Therefore, in this paper, we explore the question ``Can AI assistants know what they don’t know and express them through natural language?".

\begin{figure}[t]
\begin{center}
\centerline{\includegraphics[width=.6\linewidth]{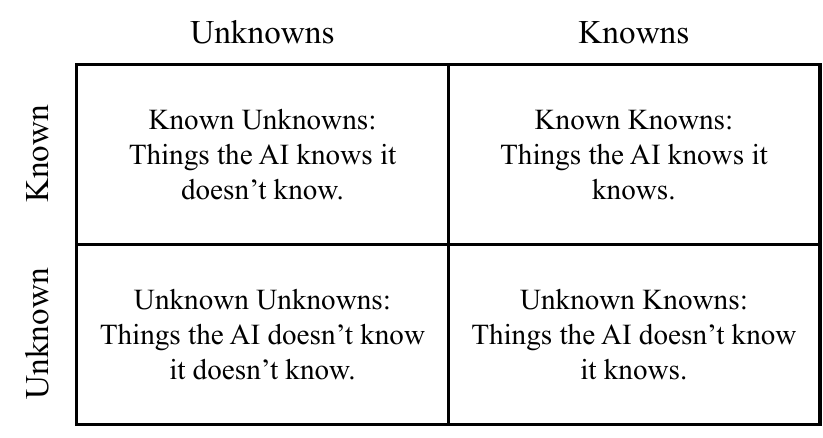}}
\caption{Knowledge quadrants of an AI assistant.
``Unknowns" represents what the AI does not actually know.
``Knowns" represents what the AI actually knows.
``Known" represents what the AI believes it knows.
``Unknown" represents what the AI believes it does not know.
}
\label{fig:Knowledge_Quadrants}
\end{center}
\end{figure}

The AI assistant's perception of its own knowledge can be represented through knowledge quadrants \citep{Self-aware}.
The knowledge quadrant is a partition which can divide the knowledge into four categories: Known Knowns, Known Unknowns, Unknown Knowns and Unknown Unknowns, as shown in Figure \ref{fig:Knowledge_Quadrants}.
Known Knowns is crucial for a truthful AI assistant, as it relies on its own knowledge to provide accurate and reliable responses.
The more knowledge that falls under the category of Known Knowns, the more helpful the AI assistant becomes.
We use \textsc{Ik-Ik} (I know I know) to represent Known Knowns.
Besides, we argue that a truthful AI assistant should also be aware of and express its lacks in certain knowledge.
Specifically, it should admit when it doesn't have information on a topic or when the information is not certain to maintain truthfulness.
This part of knowledge falls under the category of Known Unknowns.
We use \textsc{Ik-Idk} (I know I don't know) to represent Known Unknows.
Unknown Unknowns and Unknown Knowns will cause untruthful and helpless generations.
We use \textsc{Idk-Idk} (I don't know I don't know) and \textsc{Idk-Ik} (I don't know I know) to represent Unknown Unknowns and Unknown Knowns respectively.
To make AI assistants truthful, we need to teach AI assistants to know what they know and what they do not know to convert Unknown Knowns and Unknown Unknowns to Known Knowns and Known Unknowns.

\begin{figure*}[t]
    \centering
    \includegraphics[width=\linewidth]{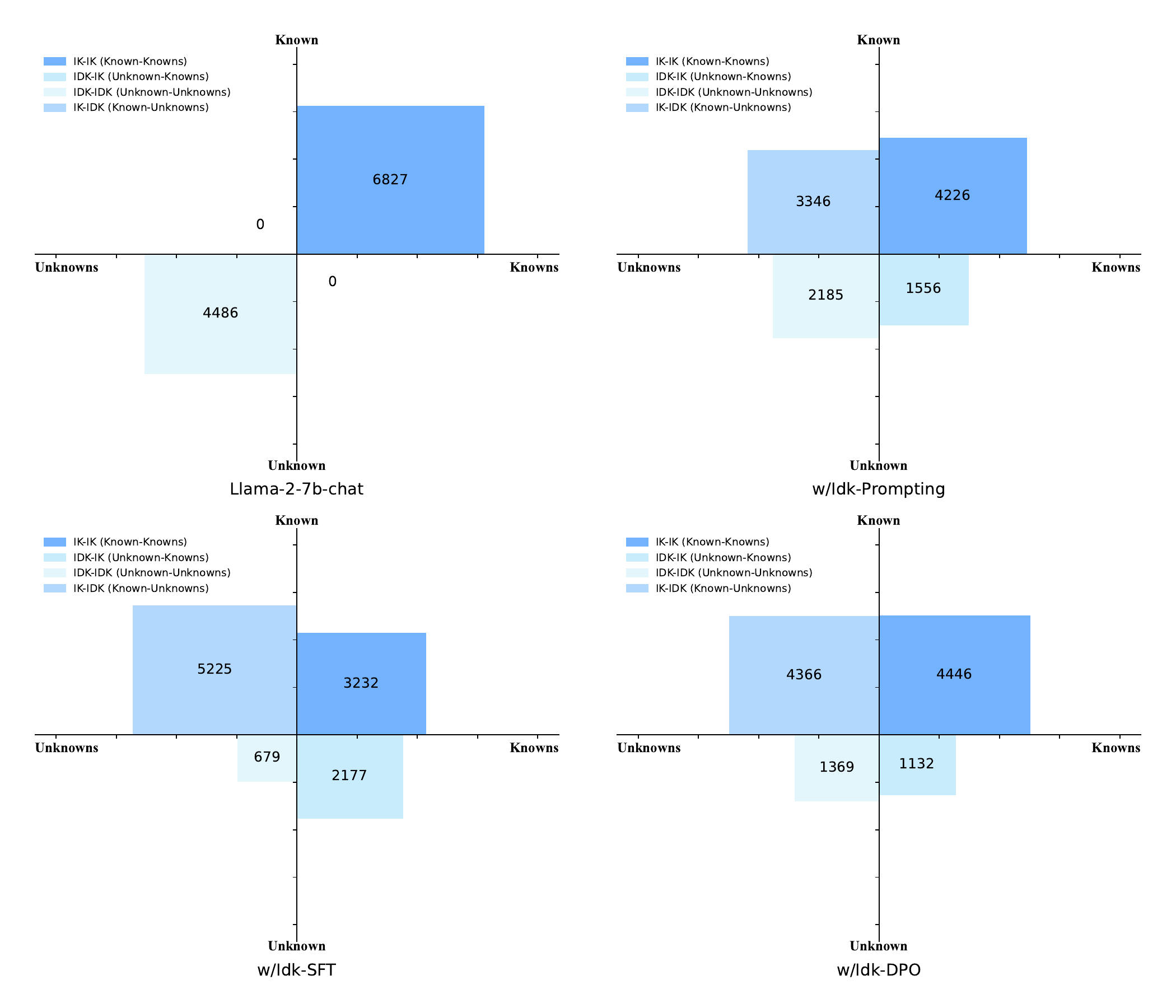}
    \caption{Knowledge quadrants of AI assistants on the Idk dataset (Ik threshold=1.0).
\textbf{\textsc{Ik-Ik}} represents the AI answers the questions correctly.
\textbf{\textsc{Idk-Ik}} represents the AI knows the answer but refuses to respond to the question.
\textbf{\textsc{Idk-Idk}} represents the AI answers the question incorrectly.
\textbf{\textsc{Ik-Idk}} represents the AI doesn't know the answer and refuses to respond to the question.
\textbf{w/Idk-Prompting:} Using prompting can transform certain \textsc{Idk-Idk} questions to \textsc{Ik-Idk} questions.
\textbf{w/Idk-SFT:} Idk-SFT allows the model to refuse to answer more questions it does not know, but it also tends to make the model more convervative, leading to incorrect refusals to answer some questions that it actually knows.
\textbf{w/Idk-DPO:} Using preference-aware optimization, like DPO, can alleviate the model's excessive conservatism and reduce the number of \textsc{Idk-Ik} questions.
}
\label{fig:knowledge_quadrants_change}
\end{figure*}

Our approach is to align an AI assistant (like llama-2-7b-chat) with a model-specific \textbf{``I don't know" (Idk)} dataset which contains the assistant's known and unknown questions.
We construct the Idk dataset based on an existing knowledge-intensive open-domain question answering dataset, TriviaQA \citep{TriviaQA}.
We determine whether an assistant knows the answer to a question by evaluating its average accuracy across multiple responses to that question.  
Questions that the assistant answers incorrectly multiple times are marked as ones it does not know, and a template for refusal to answer is annotated.
For questions that the assistant answers correctly multiple times, a correct answers it generates are used for the annotated answer.
The assistant's accuracy threshold for being considered knowing the answer to a question is a hyperparameter, which we call \textbf{Ik threshold}.
We discuss the details of constructing the Idk dataset in Section \ref{sec:idk_construction}.

In order to teach AI assistants to know what they don't know, we conduct extensive experiments to exploit the most effective method, including prompting, supervised fine-tuning and preference-aware optimization.
For prompting, we instruct the assistant to refuse answering questions it does not know through a prompt.
For supervised fine-tuning (SFT), we directly fine-tune the original assistant using our Idk datasets.
For preference-aware optimization, we use best-of-n sampling (BoN), proximal policy optimization (PPO) \citep{PPO, InstructGPT}, direct preference optimization (DPO) \citep{DPO} and hindsight instruction relabeling (HIR) \citep{HIR}.
We demonstrate some representative results in Figure \ref{fig:knowledge_quadrants_change}.

The original model (llama-2-7b-chat) can be considered as lacking the ability to recognize questions it does not know\footnote{We conducted a search for keywords such as ``I don't know", ``not sure", ``Sorry" in the responses of Llama-2-7b-chat and found that only a very small number of responses contained these keywords.}.
It may guess an answer even it lacks the knowledge.
As shown, there are many \textsc{Idk-Idk} questions, making the assistant untruthful.
Instructing the model to refuse answering unknown questions through a prompt can be effective to some extent, but there are still numerous \textsc{Idk-Ik} and \textsc{Idk-Idk} questions.
After supervised fine-tuning using Idk dataset, the number of \textsc{Idk-Ik} and \textsc{Idk-Idk} has significantly decreased, indicating that the model's ability to be aware of its own knowledge has been enhanced.
However, the model may also refuse to answer some questions it knows, leading to a decrease in the number of \textsc{Ik-Ik} questions.
Compared to SFT model, preference-aware optimization (like DPO) can mitigate the phenomenon where the model incorrectly refuses to answer questions it knows.
Besides, we conduct extensive ablation experiments to explore the effect of Ik threshold, data sources, model size and other settings.

Our findings can be summarized as follows:
\begin{itemize}
    \item[1.] After aligning using Idk datasets, AI assistants are capable of largely knowing what they know and what they do not know and refusing their unknown questions. Llama-2-7b-chat can definitively determine whether it knows the answer to up to \textbf{78.96\%} of the questions in the test set. And it exhibits good performance on out-of-distribution test sets.
    
    \item[2.] Supervised fine-tuning cause the model to become overly conservative, incorrectly rejecting known questions. Preference-aware optimization can mitigate this problem, promoting the overall proportion of \textsc{Ik-Ik} and \textsc{Ik-Idk} questions.
    
    \item[3.] The Ik threshold used to define knowns and unknowns questions influences the behavior of the assistant. The more questions labeled as "I don't know," the more likely the assistant is to refuse to answer questions. In general, the higher the Ik threshold, the greater the total number of Ik-Ik and Ik-Idk questions, resulting in a more truthful assistant.

    \item[4.] Larger model is more adept at distinguishing which questions it knows and which it doesn't know. The use of Idk-SFT on Llama-2-70b-chat, as compared to Llama-2-7b-chat, results in a 5.8\% improvement in the total number of \textsc{Ik-Ik} and \textsc{Ik-Idk} questions.

\end{itemize}

\section{Background}
\subsection{Aligning LLMs with Human Values}  
To build AI assistants based on large language models, we typically need to align these large language models with human values, making them helpful, truthful and harmless \citep{HHH, Anthropic_HH, InstructGPT}.
Here we introduce several mainstream alignment methods related to our work.
The most common alignment method for pre-trained models is instruction tuning, also known as Supervised Fine-Tuning (SFT).
\citet{flan, T0} fine-tune pre-trained models on a collection of NLP datasets combined with natural language instructions to enhance zero-shot performance on unseen tasks.
\citet{flan-t5, flanv2} scale the number of tasks, the model size and fine-tune on mixed data.
\citet{sun2023moss} utilize Self-Instruct \citep{Self-instruct} to synthesize three types of SFT data - helpful, honest, and harmless - and construct a conversational assistant.
The step following SFT is preference optimization.
\citet{Anthropic_HH, InstructGPT} use Reinforcement Learning from Human Feedback (RLHF) \citep{RLHP, Summary_RLHF}.
They first train a reward model on the human preference data and then optimize the policy model using Proximal Policy Optimization (PPO) \citep{PPO} with the trained reward model.
\citet{HIR} propose a reward-free method named Hindsight Instruction Relabeling (HIR) to utilize preference data by converting feedback to instructions and training the model using supervised fine-tuning.
\citet{DPO} propose Direct Preference Optimization (DPO) which can directly fine-tune language models to align with human preferences without the need of reward modeling.

\subsection{Discovering LLMs’ Knowledge}
Large language models store extensive world knowledge during the pre-training.
There has been increasing interest in researching knowledge in large language models.
\citet{Know_what_they_know, express_uncertainty} fine-tune language models using a classification head or verbalized confidence.
However, they don't teach models to aware their knowledge boundary and refuse to answer the questions they don't know.
\citet{Self-aware} propose the SelfAware dataset to evaluate the ability of LLMs to recognize what they don't know, finding that there is still an apparent gap compared to human.
\citet{discovering_latent_knowledge} propose an unsupervised method to find latent knowledge inside the activations of a language model by answering yes-no questions given only unlabeled model activations.
\citet{LLM-KB} investigate whether LLMs can perceive their knowledge boundaries or not under retrieval-augmented setting and normal setting.
\citet{Knowing-What-LLMs-Do-Not-Know, SelfCheckGPT, INSIDE} check the generated answers in an unsupervised way to judge whether the model knows the question.
The basic idea is that if the model knows the answer to a question, then the diversity of multiple sampling generations should be relatively low.

\subsection{Mitigating LLMs' Factual Errors}
There are some studies focused on eliminating factual errors in AI assistants.
\citet{Self-RAG} propose a framework named SELF-RAG to enhance an LM's factuality by retrieval augmentation and self-reflection.
\citet{ITI} first find truthful directions through probing and then do inference-time intervention in these truthful directions.
\citet{Representation_Engineering} use representation engineering to enhance factuality of the model's output.
\citet{Dola} propose a simple decoding strategy for reducing hallucinations by contrasting the differences in tokens' logits obtained from different layers.
\citet{Fine-tune_for_factuality} directly fine-tune language models to learn factuality from preference dataset using direct preference optimization.
However, there is currently no method that can guarantee the complete elimination of factual errors.
In practical applications, it is a necessary feature for AI assistants to refuse to answer questions they do not know.

\begin{figure*}[t]
\vskip 0.1in
\begin{center}
\centerline{\includegraphics[width=\linewidth]{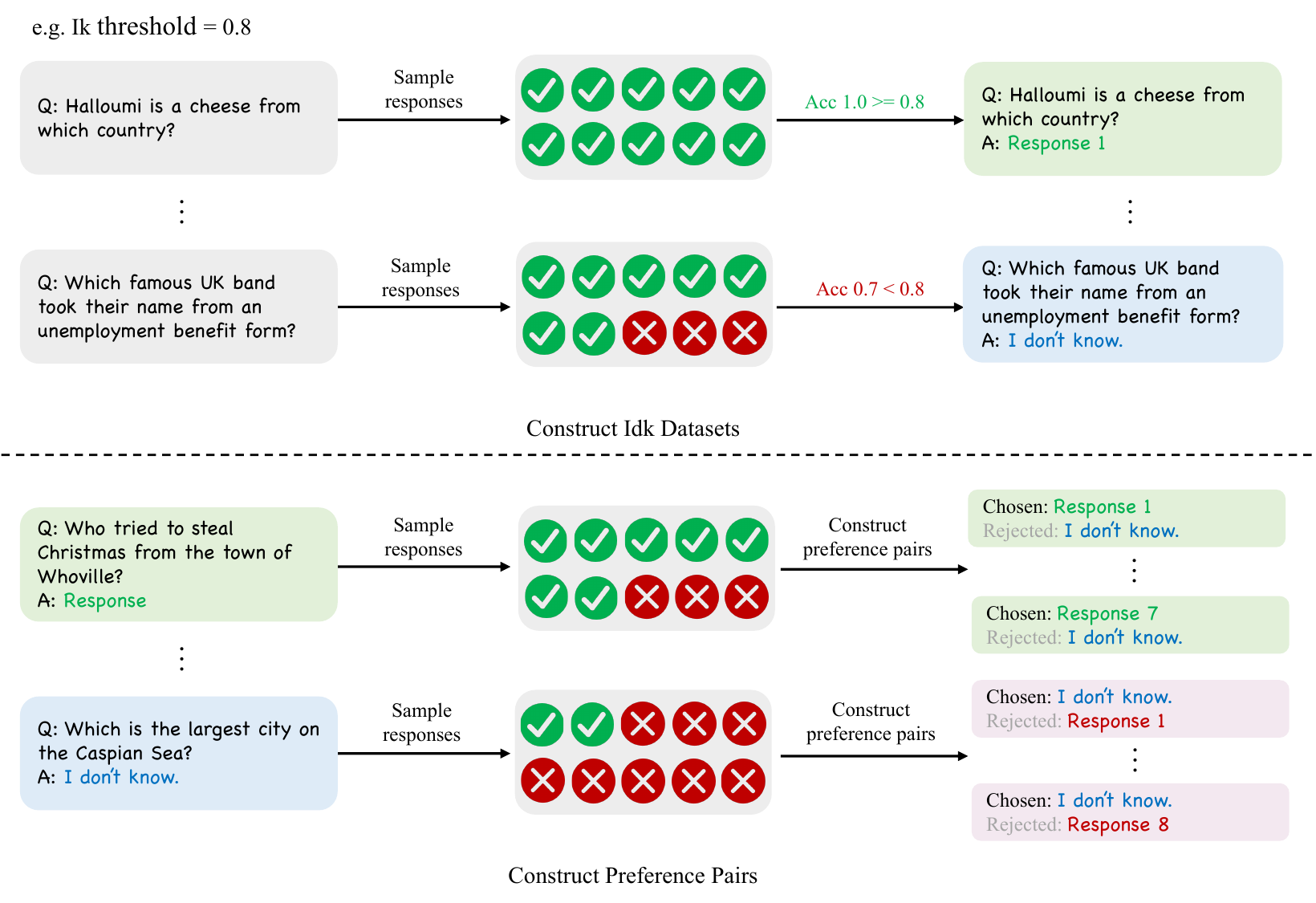}
}
\caption{\textbf{Top}: Construction process of the Idk dataset. \textbf{Bottom}: Construction process of preference pairs.
The green \textcolor{correct_response_color}{response} indicates a correct answer, the red \textcolor{wrong_response_color}{response} indicates an incorrect answer, and ``\textcolor{idk_response_color}{I don't know}" represents the template for refusal to answer.
}
\label{fig:construction_Idk_data_and_preference_pairs}
\end{center}
\end{figure*}

\section{Methodology}

\subsection{Construction of the Idk Dataset}
\label{sec:idk_construction}
Given a knowledge-intensive question answering dataset, it is hard to precisely determine which questions the model truly knows the answers to, which due to the model having varying degrees of confidence for different knowledge.
Therefore, following the approach of previous work \citep{Know_what_they_know, express_uncertainty}, we sample multiple responses from the model for each question, calculating the accuracy rate across these responses as a measure of the model's confidence regarding that question.
Finally, we select a specific level of confidence as the criterion for determining whether the model knows or does not know the answer to a question, that is the Ik threshold.
To construct the QA pairs in the Idk dataset, for questions that the model does not know, we use a template for refusal to reply as the answer.
For questions that the model knows, we select a correct response generated by the model itself as the answer.
The procedure is demonstrated in Figure \ref{fig:construction_Idk_data_and_preference_pairs} (top).
Our refusal to answer template is: 
{\footnotesize
\begin{lstlisting} [frame=none]
This question is beyond the scope of my knowledge, and I am not sure what the answer is.
\end{lstlisting}
}
We use ``I don't know" and the Idk template to refer to this template in the following paper.

\textbf{Determine whether the output of a model is correct. }
In order to construct the Idk dataset, we need an automatic method to evaluate whether the model's outputs are correct.
According to the experimental results presented in \citet{QA-eval}, employing lexical matching, which checks whether the golden answers appear in the responses generated by the model, to evaluate on a subset of the TriviaQA's validation set \citep{TriviaQA} yields a consistency rate of approximately 90\% with human evaluations.
We consider lexical matching to be a relatively accurate automatic evaluation method for the TriviaQA dataset.
Besides, TriviaQA is a mainstream knowledge-intensive open-domain question answering dataset.
Therefore, we construct our Idk dataset based on the TriviaQA dataset.

\textbf{Meaning of different Ik thresholds. }
The model is required to refuse to answer questions where the confidence level is below the Ik threshold.
It is important to note that different Ik thresholds will result in different Idk datasets.
A high Ik threshold indicates that the model will only answer a question if it possesses a high level of confidence.
Conversely, if the Ik threshold is low, the model will answer questions with a lower level of confidence required.
In other words, a high Ik threshold represents a conservative response strategy, whereas a low Ik threshold represents a more aggressive response strategy.
In this work, we sampled ten responses for each question and derived ten discrete Ik thresholds based on varying accuracy rates.
For the sake of simplicity, we set the Ik threshold to 1.0, meaning that the model is considered to know the answer to the question only if all ten of its responses are correct.
Unless specifically stated otherwise, the Idk dataset mentioned hereafter is constructed based on an Ik threshold of 1.0.
We discuss the impact of different Ik thresholds in Section \ref{sec:experiments}.

In the following sections, we introduce our methods to teach AI assistants to say ``I don't know" when encounter unknown questions.
Since the AI assistant we discuss is based on large language models, we will interchangeably use the terms ``model" and ``assistant" in the following sections.

\subsection{Idk Prompting}
For models capable of following human instructions, such as Llama-2-7b-chat, We can directly instruct an assistant to say ``I don't know" to unknown questions by adding a prompt in front of the input question.
We call this method Idk-Prompting.
This requires the model to have a high capability for following instructions, but the advantage is that it eliminates the need for additional training.
We call such a prompt an Idk prompt.
Our Idk prompt is as follows:
{\footnotesize
\begin{lstlisting} [frame=none]
Answer the following question, and if you don't know the answer, only reply with "I don't know": <Question>
\end{lstlisting}
}
As for pre-trained models lacking the ability to follow instructions, Idk-Prompting may not yield satisfactory results.

\subsection{Idk Supevised Fine-tuning}
Supervised Fine-tuing is a simple yet effective alignment method.
We directly use the Idk dataset for Supervised Fine-tuning of the model.
Since the Idk dataset contains both questions and responses, this constitutes a conditional generation task.
We input the questions into the model and require the model to predict the responses.
We perform the standard sequence-to-sequence loss to train our model.
SFT details are demonstrated in Appendix \ref{apx:SFT_details}.

\subsection{Preference-aware Optimization}
In this section, we introduce how we conduct preference-aware optimization to help the model perceive its internal knowledge better.

\paragraph{Direct Preference Optimization (DPO)}
To conduct DPO, we first train a SFT model on half of the Idk dataset as a warm up, then we collect responses from this SFT model on the other half of the Idk data.
For a given question, we conduct random sampling to gather multiple responses.
Finally, we construct preference data based on the these generated responses.
We demonstrate the procedure in Figure \ref{fig:construction_Idk_data_and_preference_pairs} (bottom).
A preference data sample consists of a question, a chosen response, and a rejected response.
The questions in the Idk dataset can be categorized into two types: those the model knows and those it does not know.
For questions the model knows, we use the correct response generated by it as the chosen response and ``I don't know" as the rejected response.
For questions the model does not know, we use ``I don't know" as the chosen response and its incorrectly generated response as the rejected response.
Besides, we find that only using the DPO loss \citet{DPO} can occasionally result in the model's inability to accurately generate the Idk template.
Therefore, in addition to the original DPO loss, we also incorporate SFT loss for the chosen responses and multiply it by a coefficient $\alpha$.
The details of the DPO are demonstrated in Appendix \ref{apx:DPO_details}.
 
\paragraph{Best-of-n Sampling (BoN)}
We also try to determine if the model knows the answer to a certain question by training a reward model to score the candidate responses.
We first train a SFT model using a half of the Idk data and then use the SFT model to initialize the reward model.
After collecting responses on the other half of the Idk dataset and constructing preference data using the same procedure as \ref{fig:construction_Idk_data_and_preference_pairs}, we train the reward model using a pairwise loss.
During inference, we employ the Best-of-10 strategy.
First, we sample ten responses using the SFT model, then we score these candidate responses with the reward model.
The response with the highest reward score is selected as the final response.
The details of reward modeling are demonstrated in Appendix \ref{apx:BoN_details}.

\paragraph{Proximal Policy Optimization (PPO)}
Based on our reward model, we can use proximal policy optimization to optimize the model.
We use the same inputs for PPO training as we do for reward modeling, but sample responses in an online manner.
The details of the PPO are demonstrated in Appendix \ref{apx:PPO_details}.

\paragraph{Hindsight Instruction Relabeling (HIR)}
So far, our Idk dataset is constructed based on a fixed Ik threshold.
In order to utilize all Idk datasets constructed with different Ik thresholds, inspired by Hindsight Instruction Relabeling \citep{HIR}, we design an instruction format to relabel all Idk datasets.
Specifically, we prepend the following instruction to each question in the Idk datasets:
{\footnotesize
\begin{lstlisting} [frame=none]
Your current knowledge expression confidence level is <X>, please answer the user's question: <Question>
\end{lstlisting}
}
where $<Question>$ is a question from an Idk dataset and $<X>$ is the value of model's knowledge confidence level ranging from 0 to 1.0, derived from the Ik threshold corresponding to the Idk dataset.
The lower the knowledge confidence level, the more inclined the model is to refuse answering questions.
Then we use the combined Idk dataset to perform supervised fine-tuning.
The advantage of using instruction relabeling is that we can control the model to adopt either a conservative or aggressive response strategy through the instruction, without the need to retrain the model.
The details of HIR are demonstrated in Appendix \ref{apx:HIR_details}.

\section{Experiments}
\label{sec:experiments}

\subsection{Dataset}
TriviaQA \citep{TriviaQA} is a reading comprehension dataset, but its question-answer pairs can be used for open-domain question answering tasks.
We use the training set of TriviaQA, consisting of 87,622 samples, to construct the training and development sets of the Idk dataset.
Since TriviaQA does not provide ground truth for the test set, we use the development set of TriviaQA to construct the test set for the Idk dataset, which comprises a total of 11,313 samples.
The detailed statistical information of the Idk datasets is provided in Appendix \ref{apx:dataset_details}.

Additionally, we use the Natural Questions (NQ) \citep{NQ} and ALCUNA \citep{ALCUNA} datasets as the out-of-distribution (OOD) questions.
NQ is a question answering dataset consisting of real queries issued to the Google search engine.
We use the development set of NQ-Open, which contains 3,610 samples, to construct the OOD test set.
According to the experimental results in \citet{QA-eval}, using lexical matching for the automatic evaluation on the development set of the NQ dataset shows more than 80\% consistency with human expert assessments.
Therefore, we use lexical matching to judge the correctness of model answers and to label ``I don't know" responses.

ALCUNA is a benchmark to assess LLMs' abilities in new knowledge understanding.
It creates new artificial entities by altering existing entity attributes and generates questions about these artificial entities.
Since these entities are artificially created, the model cannot possibly possess this knowledge.
Therefore, we use a portion of the questions from ALCUNA to test whether the model can refuse to answer, totaling 8,857 samples.

\subsection{Metrics and Evaluation}
We report the following metrics.
\begin{itemize}
\item \textsc{Ik-Ik} Rate: I know what I know (Ik-Ik) rate represents the proportion of questions answered correctly by the model out of all questions.

\item \textsc{Ik-Idk} Rate: I know what I don't know (Ik-Idk) rate represents the proportion of questions that the model correctly refuses to answer, out of all questions.
\item \textsc{Truthful} Rate: Truthful rate is the sum of Ik-Ik rate and Ik-Idk rate.
It represents the proportion of questions for which the model provides truthful responses.
The higher the value of \textsc{Truthful} rate, the clearer the model's perception of what it knows and does not know, which also indicates a higher level of truthfulness.
The \textsc{Truthful} value of an ideal truthful model should ideally reach 100\%.
\end{itemize}

The higher these three metrics are, the better.
Among these metrics, we argue that the \textsc{Truthful} rate is the most important one because it indicates the probability that users will receive a truthful response.

To calculate these metrics, we categorize the inference results into four knowledge quadrants using the following method.
\begin{itemize}
    \item \textsc{Ik-Ik}: If a question model does not refuse to answer and the answer is correct, then the question belongs to the Ik-Ik category.
    We determine whether the model's answer is correct by checking if the ground truth appears in the model's response.
    \item \textsc{Ik-Idk}: If a question model refuses to answer, and the question is marked as one that the model does not know, then this question belongs to Ik-Idk category.
    We determine whether the model refuses to answer a question by checking whether the refusal template appears in the model's response.
    \item \textsc{Idk-Ik}: If a question model refuses to answer, but the question is not marked as one the model does not know, then this question falls into the Idk-Ik category.
    \item \textsc{Idk-Idk}: If a question model does not refuse to answer but provides an incorrect response, then the question belongs to the Idk-Idk category.
\end{itemize}

We use Llama-2-7b-chat as our initial model for further training, with specific training details introduced in Appendix \ref{apx:training_details}.
We test the trained model on the test set of the Idk dataset to evaluate whether the model can distinguish between questions it knows and does not know.
Except for Idk-BoN, we use greedy decoding in all tests.
For Idk-BoN, we set the temperature coefficient to 1.0 and top\_p to 0.9, sample ten responses, and then score them using the reward model.
The response with the highest reward score is selected as the final model response.

\subsection{Main Results}

\begin{table*}[htbp]
\centering
\small
\begin{tabular}{l|ccc|ccc|c}
\toprule
& \multicolumn{3}{c|}{\textbf{TriviaQA}} & \multicolumn{3}{c|}{\textbf{Natural Questions}} & \multicolumn{1}{c}{\textbf{ALCUNA}}\\
& \textsc{Ik-Ik} & \textsc{Ik-Idk} & \textsc{Truthful} & \textsc{Ik-Ik} & \textsc{Ik-Idk} & \textsc{Truthful} & \textsc{Ik-Idk} \\ 
\midrule
Idk-Dataset$_{test}$  & 45.05 & 54.95 & 100.00 & 24.65 & 75.35 & 100.00 & 100.00 \\
\midrule
Idk-Prompting  & 37.36 & 29.58 & 66.93 & 19.75 & 41.72 & 61.47 & 91.67 \\
Idk-SFT & 28.57 & 46.19 & 74.75$_{\uparrow 7.82}$ & 15.93 & 53.99 & 69.92$_{\uparrow 8.45}$ & 98.01 \\
Idk-DPO & \textbf{39.30} & 38.59  & 77.89$_{\uparrow 10.96}$ & 20.91 & 45.60  & 66.51$_{\uparrow 5.04}$ & 98.08 \\
Idk-BoN$_{N=10}$ & 38.37 & 40.59 & \textbf{78.96}$_{\uparrow 12.03}$ & 20.55 & 47.40 & 67.95$_{\uparrow 6.48}$ & 98.32 \\
Idk-PPO & 35.90 & 40.57 & 76.47$_{\uparrow 9.54}$ & \textbf{23.13} & 42.08 & 65.21$_{\uparrow 3.47}$ & 92.66  \\
Idk-HIR & 27.36 & \textbf{48.55} & 75.91$_{\uparrow 8.98}$ & 15.40 & \textbf{56.90} & \textbf{72.30}$_{\uparrow 10.83}$ & \textbf{98.96} \\
\bottomrule
\end{tabular}
\caption{Overall results on the test set of the Idk dataset constructed based on TriviaQA and out-of-distribution test sets.}
\label{tab:overall_results}
\end{table*}

The overall results are in Table \ref{tab:overall_results}.
The Idk-Dataset used for evaluation contains 45.05\% \textsc{Ik-Ik} questions and 54.95\% \textsc{Ik-IDK} questions, which can be seen as two upper bounds of \textsc{Ik-Ik} and \textsc{Ik-Idk} rate.
Simply using an Idk prompt to let the model refuse to answer questions it doesn't know can have a certain effect, but the model's \textsc{Truthful} rate is still only 66.93\%.
The Idk-SFT can increase the \textsc{Truthful} rate ti 74.75\%, but this will result in a decrease in the \textsc{Ik-Ik} rate, which can be considered a form of ``alignment tax".
We find that preference optimization can encourage the model to answer questions, thereby mitigating the alignment tax.
DPO\footnote{We find that the DPO model, when refusing to answer questions within ALCUNA, occasionally rephrases our Idk template. Consequently, we utilize a substring of the original Idk template: ``I am not sure what the answer is" to detect whether the model refuse to answer the question.}, PPO, and BoN can all reduce the loss of \textsc{Ik-Ik} while maintaining a relatively high \textsc{Ik-Idk} rate.
Idk-BoN achieves the highest \textsc{Truthful} rate.
Idk-HIR combines all Idk datasets, which can improve \textsc{Ik-Idk} rate but help less for \textsc{Ik-Ik} rate.
However, Idk-HIR provides an switching method for Ik-threshold that does not need to retrain the model.
Overall, by aligning with the Idk dataset, we can transform \textsc{Idk-Ik} and \textsc{Idk-Idk} questions into \textsc{IK-IK} and \textsc{Ik-Idk} questions.
The model can have a clear perception of whether it knows the answers to most questions in the test set, significantly increasing truthfulness compared to before the alignment.
The additional experimental results are represented in Appendix \ref{apx:all_results}.

\paragraph{Evaluation on out-of-distribution data}
We also test whether the aligned model is capable of refusing to answer questions it does not know when encountering out-of-distribution (OOD) data.
We first construct the Idk dataset for testing based on Natural Questions, setting the Ik threshold to 1.0.
As shown in Tabel \ref{tab:overall_results}, the Idk dataset contains 24.65\% \textsc{Ik-Ik} questions and 75.35\% \textsc{Ik-Idk} questions, which means Natural Questions is more challenging than TriviaQA.
The results on Natural Questions are similar to those on TriviaQA.
The algined models show improvements in all metrics compared to using prompts.
In contrast to the results on TriviaQA, Idk-HIR achieves the highest \textsc{Truthful} rate, rather than Idk-BoN.
Furthermore, the models aligned using preference optimization methods exhibit a reduction in the \textsc{Truthful} rate compared to the Idk-SFT.
We believe this is due to the fact that preference optimization encourages the model to answer more questions.
We can observe that, compared to the Idk-SFT model, preference-optimized models have more \textsc{Ik-Ik} questions but less \textsc{Ik-Idk} questions.
In addition to this, we utilize ALCUNA to construct the Idk dataset, which only contains \textsc{Id-Idk} questions.
The results from Table \ref{tab:overall_results} indicate that the prompting method can already enable the model to refuse answering most unanswerable questions.
After alignment, the model achieves an even higher \textsc{Ik-Idk} rate.
The model aligned with TriviaQA demonstrates a high \textsc{Truthful} rate on Natural Questions and a high \textsc{Ik-Idk} rate on ALCUNA, suggesting that the model's behavior of refusing to answer unknown questions can be generalized to OOD data.

\subsection{Ablation Study}
In this section, we analyze which factors affect the model's ability to recognize questions it does not know through extensive abliation experiments.

\begin{table}[htbp]
\centering
\small
\begin{tabular}{l|cc|cc|c}
\toprule
& \textsc{Ik-Ik}$\uparrow$ & \textsc{Ik-Idk}$\uparrow$ & \textsc{Idk-Ik}$\downarrow$ & \textsc{Idk-Idk}$\downarrow$ & \textsc{Truthful}$\uparrow$ \\
\midrule
Idk-SFT$_{7b}$ & 28.57 & 46.19 & 19.24 & 6.00 & 74.75 \\
\midrule
\emph{w/Llama-2-13b-chat} & 33.92 & 41.43 & 17.45 & 7.20 & 75.35$_{\uparrow 0.60}$ \\
\emph{w/Llama-2-70b-chat} & 57.78 & 22.68 & 10.78 & 8.66 & 80.55$_{\uparrow 5.8}$ \\
\midrule
\emph{w/Idk-Mistral} & 18.35 & 50.65 & 27.68 & 3.31 & 69.00$_{\downarrow 5.75}$ \\
\emph{w/Idk-Baichuan} & 8.85 & 53.07 & 36.37 & 1.71 & 61.92$_{\downarrow 12.83}$ \\
\bottomrule
\end{tabular}
\caption{Results of ablation experiments.}
\label{tab:ablation_results}
\end{table}

\paragraph{Effect of model size}
The capabilities of LLMs are often closely related to the number of their parameters: models with a larger size tend to be more powerful.
We conduct Idk-SFT on Llama-2-7b-chat, Llama-2-13b-chat and Llama-2-70b-chat to observe whether the size of the model affects the effectiveness of Idk-SFT.
In Table \ref{tab:ablation_results}, we report the proportions of each knowledge quadrant for models of different sizes.
However, the label distribution of the Idk dataset corresponding to different initial models is inconsistent (the larger the model, the more \textsc{Ik-Ik} questions), as shown in Appendix \ref{apx:label_distribution}.
This results in the \textsc{Ik-Ik} rate and \textsc{Ik-Idk} rate being incomparable.
Therefore, we mainly focus on the \textsc{Truthful} rate of different models.
The \textsc{Truthful} rate of the 13B model is slightly higher than that of the 7B model.
The \textsc{Truthful} rate of the 70B model is significantly higher than that of the 13B and 7B models.
This indicates that larger models are more adpet at distinguishing between questions they know and do not know.

\paragraph{Effect of data sources}
Due to the differences in the pre-training process, different pre-trained models possess distinct knowledge.
During training, we construct model-specific Idk dataset for different pre-trained models.
This is because we want the model to determine whether it knows the answer to a question based on its internal knowledge, rather than learning to recognize some specific patterns of questions.
The model-specific Idk dataset can connect the model's internal knowledge with the labels of the Idk dataset.
To explore the impact of using a non-model-specific Idk dataset on training, we construct two Idk training sets using Mistral-7B-Instruct-v0.1 \citep{Mistral-7b} and Baichuan2-7B-chat \citep{baichuan2} respectively, named ``Idk-Mistral" and ``Idk-Baichuan".
We present label distributions of these Idk datasets in Appendix \ref{apx:label_distribution}
As shown in Tabel \ref{tab:ablation_results}, using non-model-specific Idk datasets like ``Idk-Mistral" or ``Idk-Baichuan" does result in a \textsc{Truthful} rate loss.
Due to the numerous Idk questions in the Idk-Mistral and Idk-Baichuan datasets, the trained model tends to be more inclined towards refusing to answer questions, which has resulted in a significant reduction in Ik-Ik related queries, far below their proportion in the dataset.
This indicates that constructing a model-specific Idk dataset is necessary for enabling the model to learn to refuse to answer questions it does not know.

\begin{figure}[tbp]
    \centering
    \includegraphics[width=.9\linewidth]{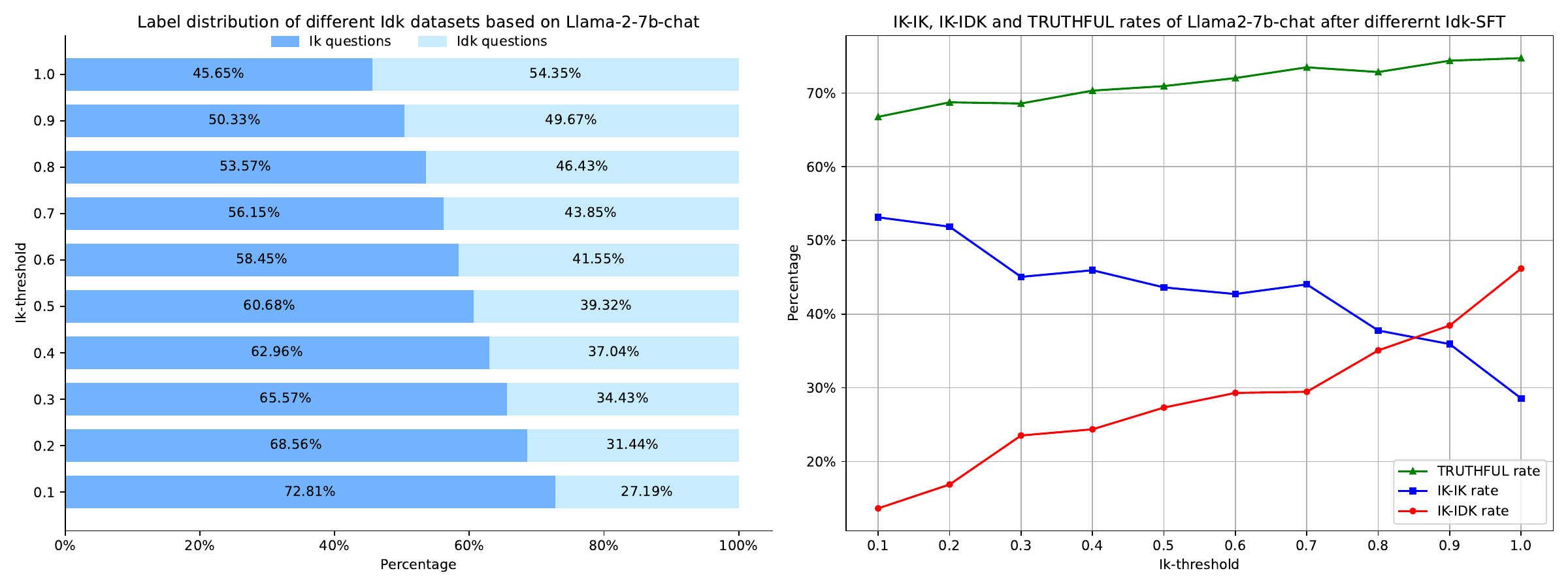}
    \caption{\textbf{Left:} Variation in the proportions of Ik and Idk questions within the Idk datasets constructed based on different Ik thresholds. \textbf{Right:} The changes in \textsc{Ik-Ik} rate, \textsc{Ik-Idk} rate, and \textsc{Truthful} rate after conducting Idk-SFT with different Idk datasets.}
    \label{fig:label_distribution_and_f1}
\end{figure}

\paragraph{Effect of IK threshold}
So far, we have fixed the Ik threshold to 1.0.
Here, we discuss the impact of different Ik thresholds on model behaviors.
The Ik threshold primarily affects the distribution of labels in the Idk dataset, with a higher Ik threshold indicating that more questions will be labeled as ``I don't know".
As demonstrated in Figure \ref{fig:label_distribution_and_f1} (left), the higher the value of the Ik threshold, the greater the proportion of Idk questions.
This is because when the Ik threshold is high, only questions with a high confidence level will be annotated as questions known to the model.
As shown in Figure \ref{fig:label_distribution_and_f1} (right), increasing the Ik threshold results in a decrease in the \textsc{Ik-Ik} rate and an increase in the \textsc{Ik-Idk} rate.
As the Ik threshold is raised, the model's \textsc{TRUTHFUL} rate will continue to improve.
In other words, setting a high Ik threshold aids the model in better distinguish between knowledge it knows and does not know, making the model more truthful.
In contrast, setting a low Ik threshold can make the model more helpful, since the number of \textsc{Ik-Ik} questions will increase.
Besides, we find that as the proportion of Idk questions in the dataset increases, the model tends to refuse to answer questions more frequently.
We report the F1 scores of Idk and Ik questions in different Idk datasets in Appendix \ref{apx:f1_results_effect_of_Ik_threshold} and the knowledge quadrants under different Ik thresholds in \ref{apx:all_knowledge_quadrants}.

\section{Conclusion}
In this paper, we explore the question ``Can AI assistants know what they don’t know?".
We find that after aligning the AI assistant with its own Idk(``I don't know") dataset which contains its known and unknown questions, the assistant can be aware of what it does not know to a certain extent.
In the given test set for open-domain question-answering, Llama-2-7b-chat can explicitly indicate whether it knows or does not know the answers to up to 78.96\% of the questions and refuse to answer the questions it does not know.
To achieve this, we utilize various methods to use the Idk dataset for alignment, including prompting, supervised fine-tuning and preference-aware optimization.
We also find that the Ik threshold influences the model's tendency to decline responses.
Empolying an Idk dataset from different models for alignment results in a performance degradation.
Furthermore, a large model like Llama-2-70b-chat achieves a higher \textsc{Truthful} rate.
An AI assistant's refusal to answer questions beyond its knowledge can reduce hallucinations.
We believe this is an essential behavior for a truthful AI assistant.

\bibliography{colm2024_conference}
\bibliographystyle{colm2024_conference}

\newpage
\appendix
\section{Idk Dataset Construction Details}
\label{apx:dataset_details}
\subsection{Data Statistics}
We use the training set of TriviaQA, consisting of 87,622 samples, to construct the training and development sets of the Idk dataset.
We partition 10\% of the training set of TriviaQA to serve as the validation set of the Idk dataset, with the other 90\% as the training set.
Therefore, the validation set contains 8,763 samples and the training set contains 78,899 samples.
We use the development set of TriviaQA to construct the test set for the Idk dataset, which comprises a total of 11,313 samples.
The number of samples in each part of the Idk dataset for different models is the same, it is only the distribution of the labels that varies.

\subsection{Sampling Parameters}
When constructing the Idk dataset through sampling model responses, our sampling parameters are set as follows: top\_p=0.9, temperature=1.0, max\_new\_tokens=512, repetition\_penalty=1.0 (no penalty).
We use this set of parameters for all random sampling in this work.

\subsection{Label Distribution of Idk Datasets}
\label{apx:label_distribution}
\begin{figure*}[ht]
    \vskip -0.1in
    \centering
    \includegraphics[width=.8\linewidth]{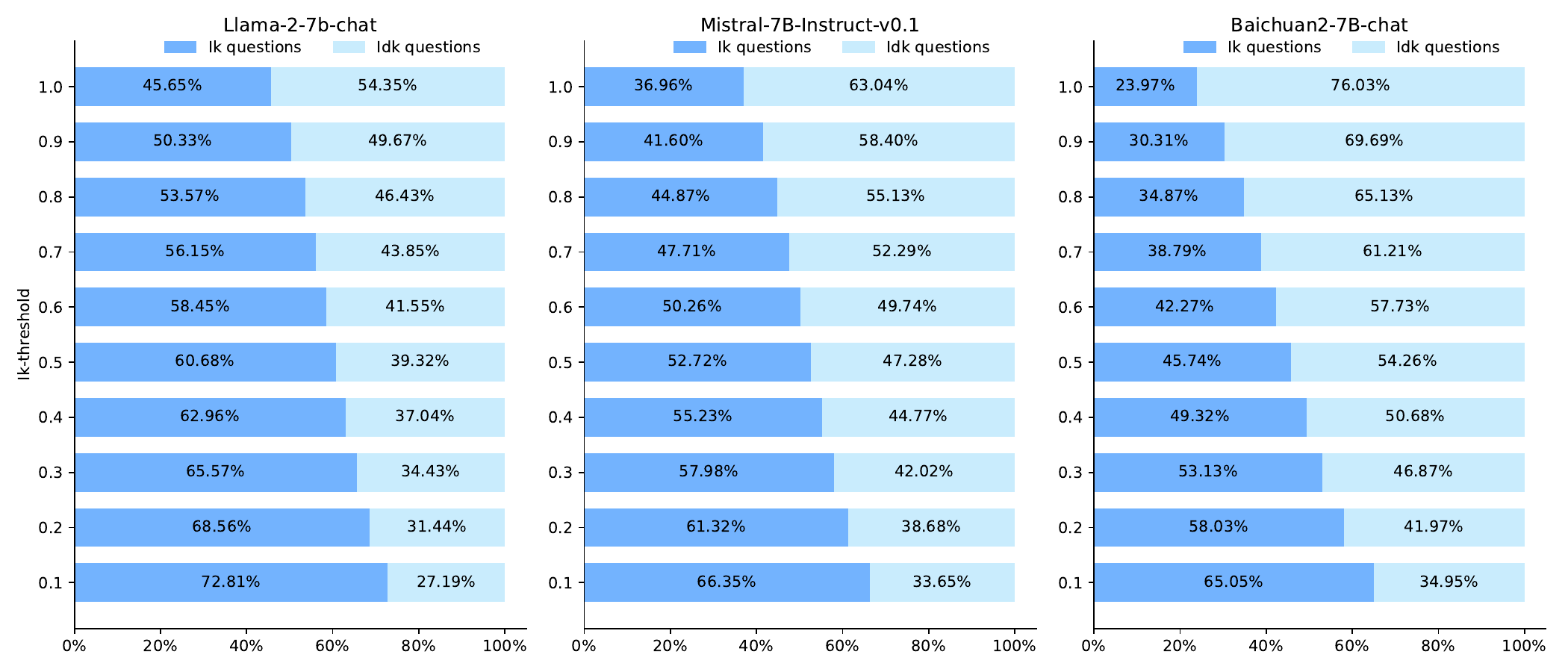}
    \caption{Label distribution in the Idk dataset across different models.}
    \label{fig:label_distribution_all}
\end{figure*}

\begin{figure*}[ht]
    \vskip -0.1in
    \centering
    \includegraphics[width=.8\linewidth]{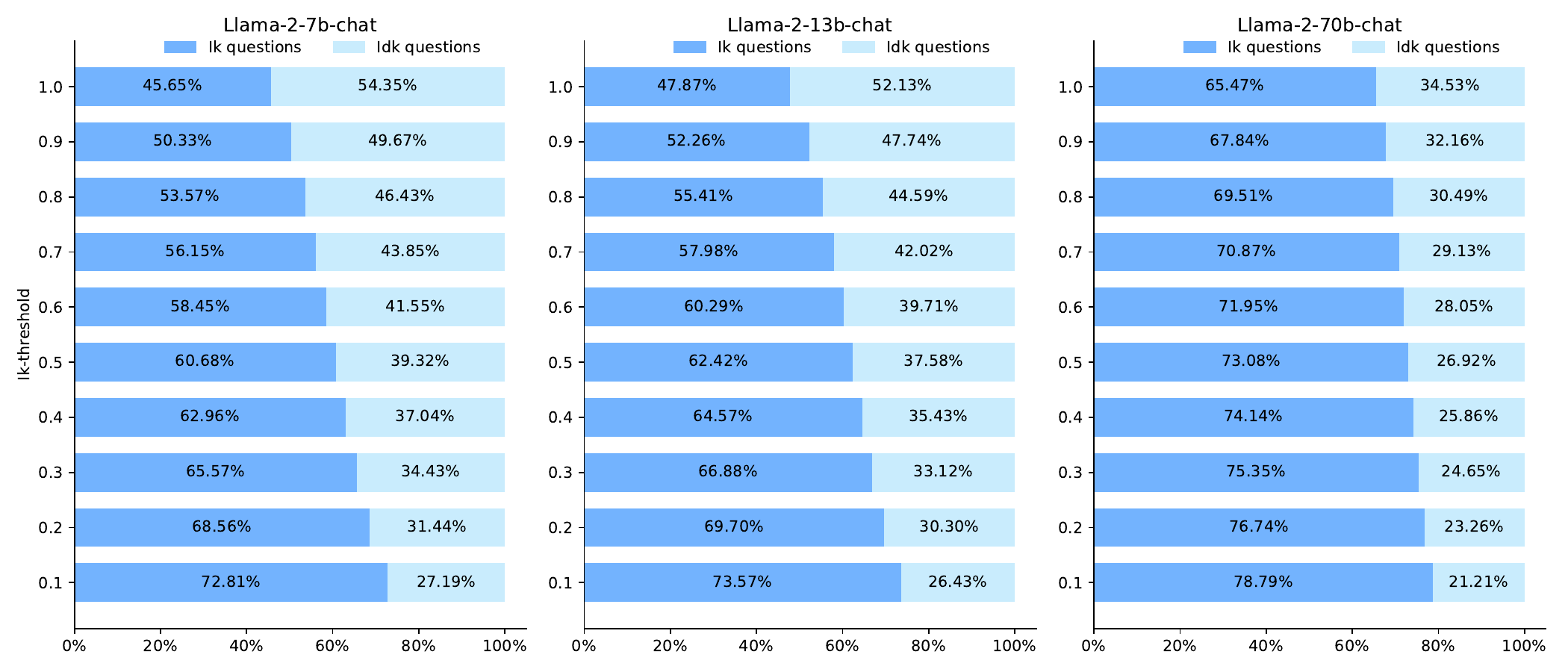}
    \caption{Label distribution in the Idk dataset across different sizes.}
    \label{fig:label_distribution_different_sizes}
\end{figure*}

In Figure \ref{fig:label_distribution_all} and Figure \ref{fig:label_distribution_different_sizes}, we present the label distribution in the Idk datasets constructed using different Ik thresholds across various models.
It is evident that different models possess varying knowledge reserves, as indicated by the distinct differences in the label distribution of their Idk datasets.
As shown in Figure 6, the larger the size of the model, the more extensive its knowledge, resulting in fewer questions being labeled as ``I don't know".

\newpage

\section{Training Details}
\label{apx:training_details}

\subsection{Supervised Fine-tuning}
\label{apx:SFT_details}
We organize our Idk dataset into single-turn dialogues following the conversation format of Llama-2-7b-chat and then use the standard SFT loss to train the model:
\begin{gather}
\mathcal{L}_{SFT} = -E_{(x,y) \sim D}[\frac{1}{N}\sum^{N}_{t} \log p(y_{t} | x, y_{<t}; \theta)]
\end{gather}
$(x, y)$ is a question-answering pair in the Idk dataset, where $x$ represents the question, and $y$ represents the answer.
N represents the length of the answer $y$, and $\theta$ represents the model parameters.
During training, we employ a packing strategy to combine multiple samples into a single sequence with a maximum length of 4096.
Following the settings of llama-recipes\footnote{\url{https://github.com/facebookresearch/llama-recipes}}, our batch size is set to 32, with a learning rate of 1e-4 and train 10 epochs.
During training, we save a checkpoint at the end of each epoch, and select the checkpoint that performs the best on the validation set as the final model.
We employed Fully Sharded Data Parallelism (FSDP) to conduct SFT training on eight A100 80G GPUs.
For Llama-2-70b-chat, we train 10 epochs using 32 A100 80G GPUs and select the checkpoint of the last epoch as the final model.
The decision to forego the use of a validation set for model selection was based on our observation that the model exhibiting the lowest loss on the validation set tended to erroneously reject numerous Ik questions.
We speculate that this may be attributed to the inherent alignment training of the Llama-2-70b-chat itself.

\subsection{Direct Preference Optimization}
\label{apx:DPO_details}
The original DPO loss proposed by \citet{DPO} is:
\begin{gather}
\mathcal{L}_{DPO} = -\mathbb{E}_{(x, y_w, y_l)\sim \mathcal{D}}\left[\log \sigma \left(\beta \log \frac{\pi_{\theta}(y_w\mid x)}{\piref(y_w\mid x)} - \beta \log \frac{\pi_{\theta}(y_l\mid x)}{\piref(y_l\mid x)}\right)\right].
\end{gather}
where $\piref$ is the SFT model trained with half of the Idk data, $\pi_{\theta}$ is the policy model, $y_w$ is the chosen response and $y_l$ if the rejected response.
To alleviate the problem of the DPO model sometimes failing to fully generate the Idk template, we additionally incorporate the SFT loss.
Our final loss function of direct preference optimization is:
\begin{gather}
\mathcal{L}_{DPO-SFT} = \mathcal{L}_{DPO} + \alpha * \mathcal{L}_{SFT}
\end{gather}
In the experiment, we set the coefficient $\alpha$ of the SFT loss to 0.01.
The hyperparameters of our SFT model training are the same as Appendix \ref{apx:SFT_details}.
During DPO training, following DPO's official implementation\footnote{\url{https://github.com/eric-mitchell/direct-preference-optimization}}, we set our batch size to 64, the learning rate to 5e-7, $\beta$ to 0.1 and train for one epoch.
We partition 10\% of the preference data to construct a validation set to select the best checkpoint.
We use 8 A100 80G GPUs for DPO training.
We present the impact of different $\alpha$ values on the model's \textsc{Truthful} rate in Table \ref{tab:effect_of_alpha}.

\begin{table}[htbp]
\centering
\caption{The impact of the coefficient $\alpha$ of the SFT loss on the model's \textsc{Truthful} rate.}
\vskip 0.1in
\small
\begin{tabular}{lccccc}
\toprule
& $\alpha=0$ & $\alpha=0.01$ & $\alpha=0.1$ & $\alpha=0.5$ & $\alpha=1.0$ \\ 
\midrule
Ik-threshold=0.5 & 74.28 & 72.39 & 72.06 & 72.31 & 72.08 \\
Ik-threshold=1.0 & 66.14 & 77.89 & 76.68 & 75.55 & 75.72 \\
\bottomrule
\end{tabular}
\label{tab:effect_of_alpha} 
\vspace{-5pt}
\end{table}

As shown in Table 4, when using the Idk dataset constructed with Ik-threshold=0.5 for DPO training, the model is capable of accurately generating the Idk template.
In this scenario, incorporating SFT loss reduces the model's \textsc{Truthful} rate.
However, when using the Idk dataset constructed with Ik-threshold=1.0 for DPO training, the model occasionally fails to accurately generate the Idk template.
In such cases, employing a coefficient of 0.01 yields the most effective mitigation.

\subsection{Best-of-n Sampling}
\label{apx:BoN_details}
We train the reward model using a pairwise loss:
\begin{gather}
\mathcal{L}_{RM} = -E_{(x,y_w,y_l) \sim D}\left[\log\sigma\left(r(x_i,y_w)-r(x_i,y_l)\right)\right]
\end{gather}
where $(x, y_w, y_l)$ is a question-chosen-rejected triplet from the preference dataset.
During training of the reward model, we set batch size to 128, learning rate to 9e-6, and train for one epoch.
We partition 10\% of the preference data to construct a validation set to select the best checkpoint.
We use 4 A100 80G GPUs for reward model training.

\subsection{Proximal Policy Optimization}
\label{apx:PPO_details}
We employ the SFT model and reward model obtained from \ref{apx:BoN_details} fro PPO training.
We use DeepSpeed-Chat \footnote{\url{https://github.com/microsoft/DeepSpeedExamples/tree/master/applications/DeepSpeed-Chat}} for PPO training.
The SFT model and reward model used in PPO training are obtained from the BoN's supervised fine-tuning and reward modeling.
For PPO \citep{PPO}, the loss function of the actor model is:
\begin{gather}
\mathcal{L}_{PPO-Actor} = -\hat{E_{t}}[max(r_t(\theta)\hat{A_t}, clip(r_t(\theta), 1-\epsilon, 1+\epsilon)\hat{A_t}],
r_t(\theta) = \frac{\pi_{\theta}(a_t|s_t)}{\pi_{\theta_{old}}(a_t|s_t)}
\end{gather}
And the loss function of the critic model is:
\begin{gather}
\mathcal{L}_{PPO-Critic} = 0.5*\hat{E_{t}}[max((V_{\phi}(s_t)-\hat{R_t})^2, clip(V_{\phi}(s_t), V_{old}(s_t)+\epsilon, V_{old}(s_t)-\epsilon))]
\end{gather}
We set the learning rate for both the actor model and the critic model to 1e-6.
The generation batch size is 64 and the training batch size is 32.
Each training step, we train a single inner epoch.
We utilize DeepSpeed ZeRO-3 to train one epoch on 32 A100 80G GPUs.

\subsection{Hindsight Instruction Relabeling}
\label{apx:HIR_details}
We combine 10 Idk datasets using the HIR method, constructed from 10 distinct Ik thresholds ranging from 0.1 to 1.0.
These Ik thresholds correspond to knowledge confidence level from 1.0 to 0.1, respectively.
The lower the knowledge confidence level, the less confident the model is in its own knowledge, resulting in a more conservative response strategy.
Besides, we also add a dataset consisting entirely of refusals to respond, corresponding to situations where the knowledge confidence level is 0 and its Ik threshold can be seen as 1.1.
We utilize the following formula to convert from the Ik threshold to the knowledge confidence level:
\begin{gather}
Knowledge\_confidence\_level = 1.1 - Ik\_threshold 
\end{gather}
For example, we prepend the following instruction to questions in the Idk dataset corresponding to an Ik threshold of 1.0:
{\footnotesize
\begin{lstlisting} [frame=none]
Your current knowledge expression confidence level is 0.1, please answer the user's question: <Question>
\end{lstlisting}
}
We set the batch size to 256, the learning rate to 2e-5 and we train for 3 epochs using 8 A100 80G GPUs.
The advantage of this method is that it allows users to control the model's response strategy without the need to retrain the model.
For instance, in scenarios where there is a low tolerance for factual errors, we can set the knowledge confidence level to 0.1.
This setting prompts the model to answer only those questions it is particularly certain about, thereby ensuring truthfulness.
Conversely, in situations where there is a higher tolerance for factual errors, we can adjust the knowledge confidence level to 1.0.
This adjustment encourages the model to respond to a wider range of questions, enhancing its helpfulness.
We show the comparison between Idk-HIR and Idk-SFT in Appendix \ref{apx:Idk-HIR_vs_Idk-SFT}.

\newpage

\section{Additional Experimental Results}
\label{apx:all_results}

\subsection{Knowledge Quadrants Under Different Ik Thresholds}
\label{apx:all_knowledge_quadrants}

\begin{figure}[htbp]
    \centering
    \includegraphics[width=\linewidth]{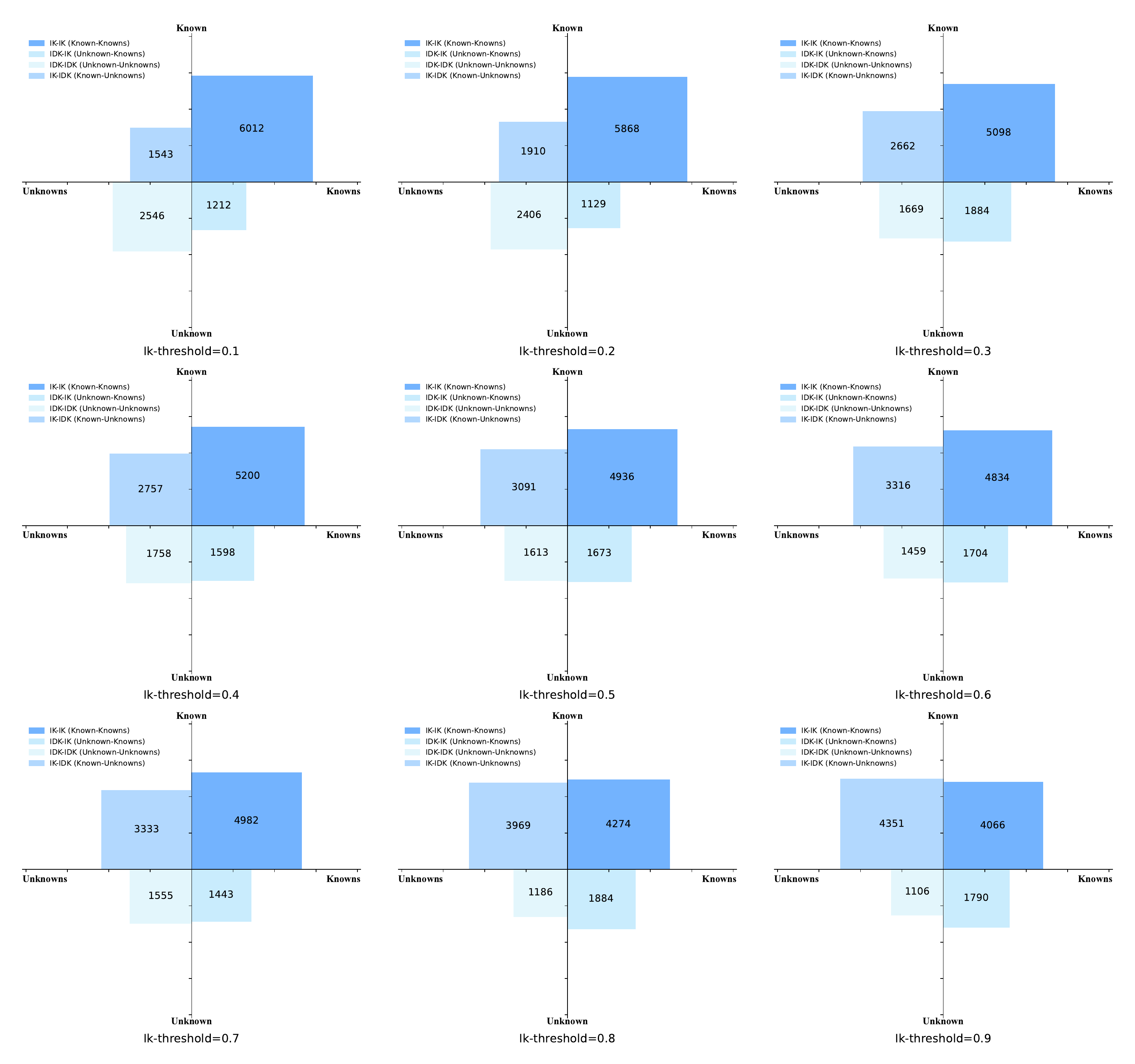}
    \caption{Knowledge quadrants under different Ik thresholds.}
    \label{fig:all_knowledge_quadrants}
\end{figure}

In Figure \ref{fig:all_knowledge_quadrants}, we present the distribution of the model's knowledge quadrants after Idk-SFT when the Ik threshold ranges from 0.1 to 0.9.

\subsection{Effect of IK threshold}
\label{apx:f1_results_effect_of_Ik_threshold}
\paragraph{Answer F1 and Refusal F1.}
We report Answer F1 score and Refusal F1 score to reflect changes in the model's behavior influenced by the Ik threshold.
Regarding Answer F1, we only consider whether the model answer the question, without taking into account the accuracy of the answer.

\begin{figure}[ht]
    \centering
    \includegraphics[width=.5\linewidth]{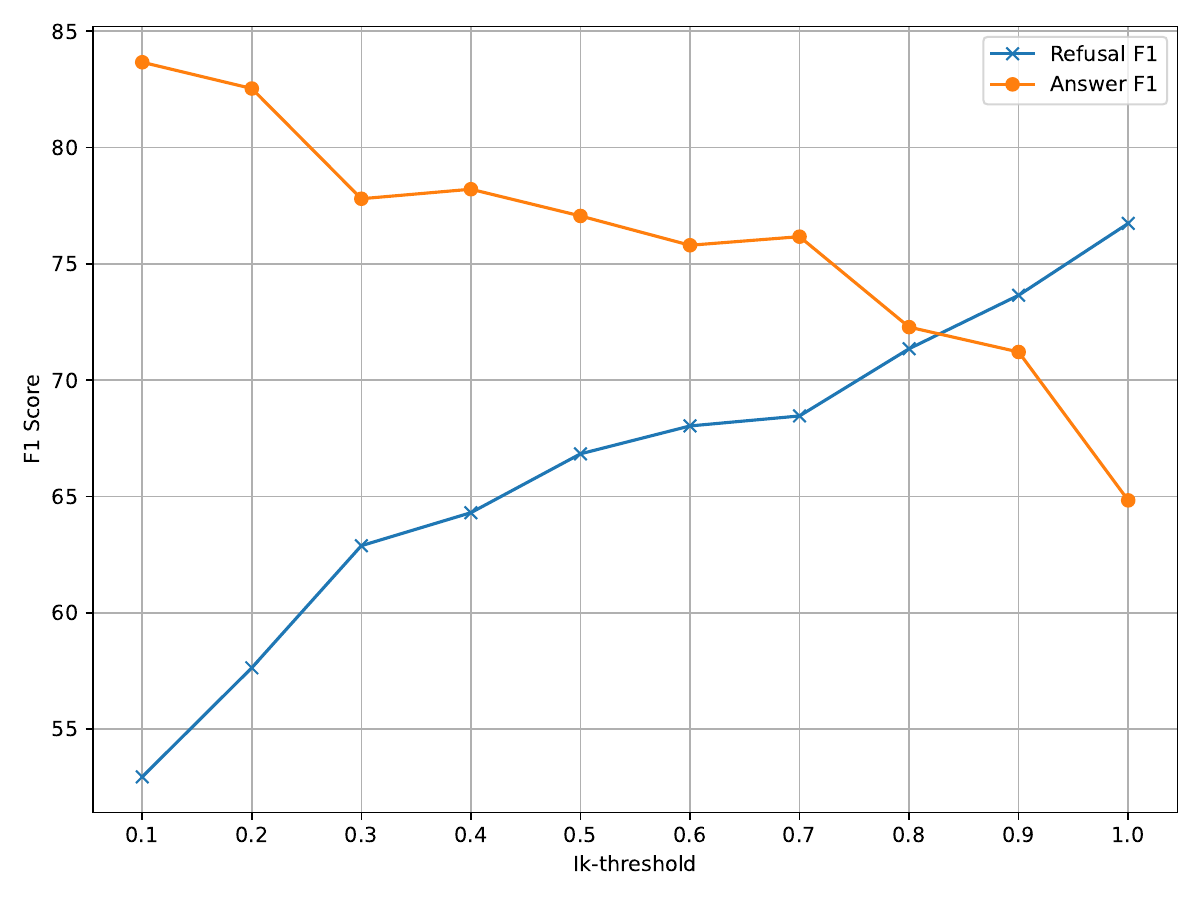}
    \caption{Refusal F1 and Answer F1 scores at different Ik thresholds.}
    \label{fig:Refusal_F1_Answer_F1}
\end{figure}

As shown in Figure \ref{fig:Refusal_F1_Answer_F1}, when the Ik threshold raises, the model tends to refuse to answer questions, resulting in an increase in Refusal F1.
Conversely, when the Ik threshold is low, the model in more inclined to answer questions, leading to an increase in Answer F1.

\subsection{Idk-HIR vs Idk-SFT}
\label{apx:Idk-HIR_vs_Idk-SFT}
In this section, we compare the effects of Idk-HIR and Idk-SFT.

\begin{figure*}[htbp]
    \centering
    \includegraphics[width=\linewidth]{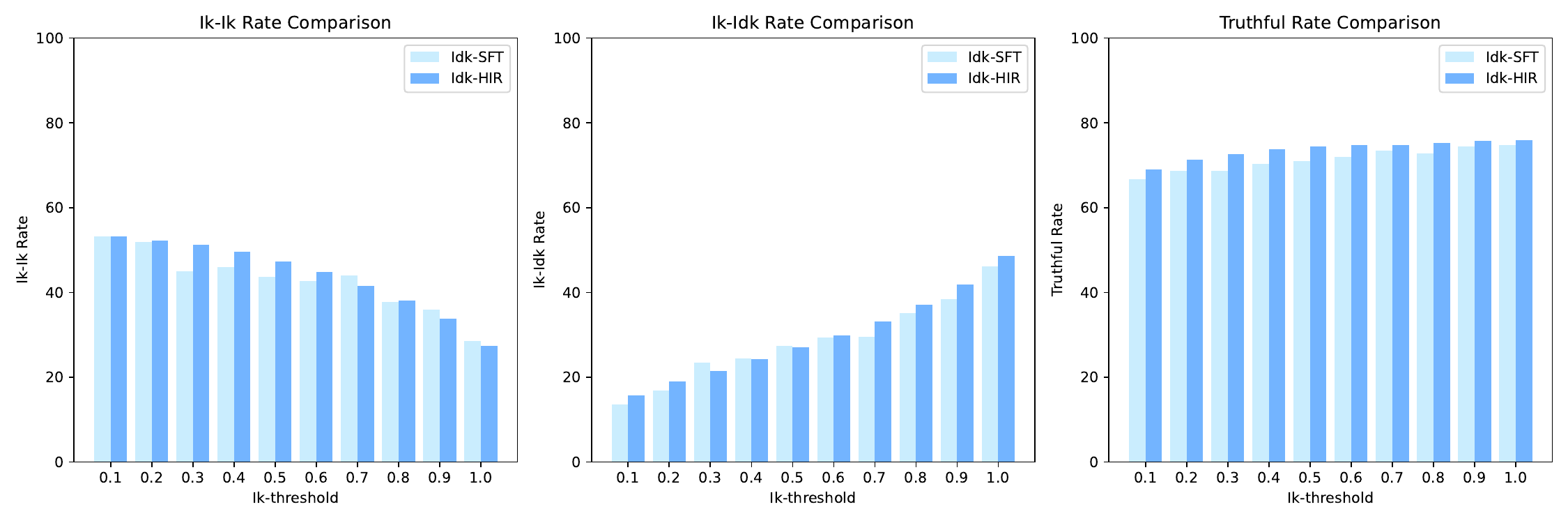}
    \caption{Comparison between Idk-SFT and Idk-HIR.}
    \label{fig:HIR_vs_SFT}
\end{figure*}

As shown in Figure \ref{fig:HIR_vs_SFT}, the \textsc{Ik-Ik} rate and \textsc{Ik-Idk} rate of the Idk-HIR model are comparable to those of the Idk-SFT model across various Ik thresholds, and the \textsc{Truthful} rate is consistently higher than that of the Idk-SFT.
Therefore, in certain scenarios, the flexible and controllable Idk-HIR model serves as an excellent alternative to the Idk-SFT model.

\end{document}